\documentclass[letterpaper,journal]{IEEEtran}
\usepackage{amsmath,amsfonts}
\usepackage{algorithmic}
\usepackage{algorithm}
\usepackage{array}
\usepackage[caption=false,font=normalsize,labelfont=sf,textfont=sf]{subfig}
\usepackage{textcomp}
\usepackage{stfloats}
\usepackage{url}
\usepackage{verbatim}
\usepackage{graphicx}
\usepackage{cite}
\usepackage{amsmath}
\usepackage{amssymb}
\usepackage{color}
\usepackage{url}
\usepackage{booktabs}
\usepackage{multirow}
\usepackage{makecell}

\begin{document}

\title{HAMP-LIC: Hessian-Aware Mixed-Precision Post-Training Quantization for Learned Image Compression}

\author{Yuefeng Zhang
\thanks{Yuefeng Zhang is with the Beijing Institute of Computer Technology and Application, Beijing, China. 
She is currently pursuing a Ph.D. degree at the School of Elite Engineering, Northwestern Polytechnical University, Shaanxi, China (e-mail: yuefeng.zhang@mail.nwpu.edu.cn).}
}

\maketitle

\begin{center}
\footnotesize
This work has been submitted to the IEEE for possible publication.
Copyright may be transferred without notice, after which this version
may no longer be accessible.
\end{center}

\begin{abstract}
Learned image compression (LIC) models achieve strong rate–distortion performance
but are hindered by high computational complexity and encoding–decoding mismatches across heterogeneous hardware platforms.
Uniform fixed-precision quantization alleviates these issues but suffers severe quality degradation at low bit-widths,
as it ignores the differing quantization sensitivities of individual layers.
To enable efficient and accurate low-bit deployment of pre-trained LIC models,
we propose HAMP-LIC, a Hessian-aware mixed-precision post-training quantization (PTQ) framework with a four-stage optimization strategy.
First, block-wise sensitivity is estimated from the Hessian trace to capture second-order importance.
Second, a task-aware refinement module adjusts these sensitivities by jointly considering quantization distortion and rate–distortion performance.
Third, guided by the refined sensitivity profile, bit-widths are allocated under a global model-size constraint to balance efficiency and reconstruction quality.
Finally, block-wise reconstruction on a small calibration set further suppresses quantization error.
Experiments on representative LIC models, including \textit{Minnen2018} and \textit{Cheng2020},
demonstrate that HAMP-LIC achieves up to 4.85$\times$ model compression with as low as 0.59\% BD-rate loss,
consistently outperforming existing fixed- and mixed-precision PTQ methods across multiple datasets
while completely eliminating cross-platform encoding–decoding errors.
\end{abstract}


\begin{figure}[t]
\begin{center}
  \includegraphics[width=0.95\linewidth]{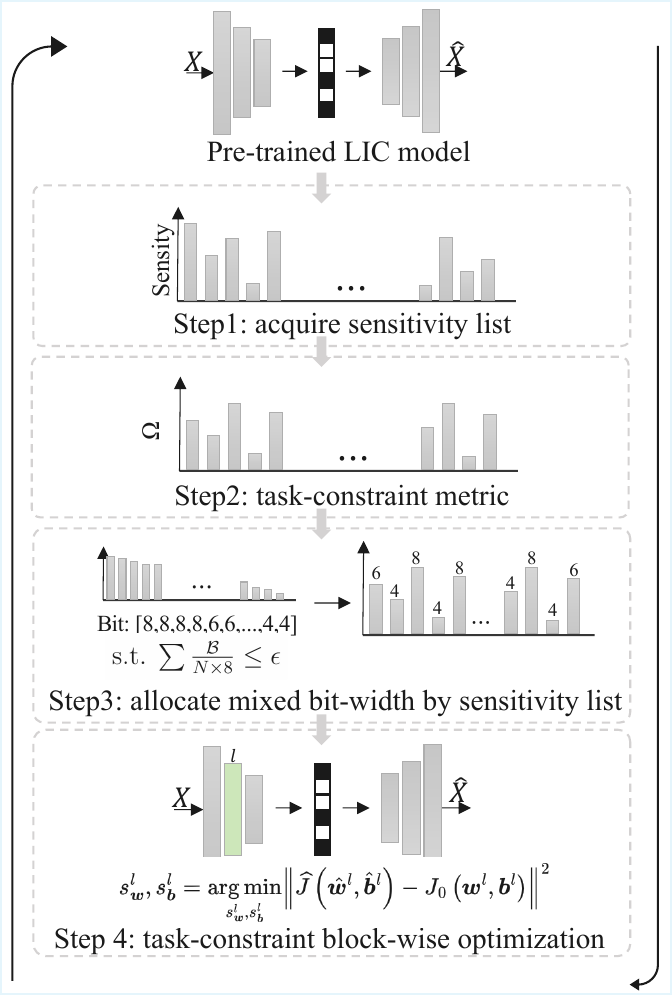}
\end{center}
\caption{The overview of the proposed quantization framework HAMP-LIC for the pre-trained LIC model, where 4 steps are conducted.}
\label{fig:overview}
\end{figure}

\section{Introduction}
Recent advances in learned image compression (LIC) have established it as a leading paradigm in image and video coding, achieving significant improvements in rate–distortion performance over traditional handcrafted codecs. 
By jointly optimizing nonlinear transforms and probabilistic entropy models, LIC systems are able to learn compact latent representations that substantially enhance compression efficiency while preserving reconstruction quality.
Despite these advantages, the computational and memory requirements of modern LIC architectures remain prohibitive for deployment on resource-constrained devices, limiting their practical applicability in real-world scenarios.

In addition to high inference cost, LIC models are also sensitive to numerical precision \cite{he2022post}. In practice, different hardware platforms exhibit inconsistent behavior in floating-point arithmetic, particularly in entropy modeling and probability estimation. Such discrepancies can lead to encoding–decoding mismatches, resulting in degraded reconstruction quality or even instability across platforms. This problem is far from negligible in practice: as shown in our experiments (Section~\ref{sec:enc_dec_error_rate}), a full-precision \textit{Cheng2020} model fails to decode up to 24 out of 24 Kodak images and 98 out of 100 Tecnick images when encoding and decoding are performed on different devices (CPU and GPU). These issues highlight the need for efficient and numerically robust compression strategies that maintain performance under low-precision computation.

To address these challenges, model quantization has emerged as a promising solution for reducing computational complexity and memory footprint. Existing quantization methods are generally categorized into Quantization-Aware Training (QAT)~\cite{balle2019integer} and 
Post-Training Quantization (PTQ)~\cite{Hong0LPM21,He2022ptq,ShiLM24,FaisalHossain2024FlexibleMP}. 
QAT incorporates quantization effects during training to improve robustness but requires full access to training data and substantial retraining cost. 
By treating all modules with a fixed precision, these methods ignore the heterogeneous sensitivity of different modules, which often leads to suboptimal rate–distortion performance under aggressive quantization. 
Although the recent FMPQ~\cite{FaisalHossain2024FlexibleMP} explores mixed-precision quantization for LIC, 
it remains limited to relatively high precision (e.g., 8-bit average) and does not fully investigate ultra-low-bit regimes.
In contrast, PTQ directly converts a pre-trained FP32 model into a low-bit representation using a small calibration set, eliminating training time. 
However, most existing PTQ approaches for LIC adopt uniform bit-width assignment across all parameters: 
Hong \emph{et al.}~\cite{Hong0LPM21} restrict fixed-point inference to the decoder with range-adaptive quantization (RAQ), 
He \emph{et al.}~\cite{He2022ptq} first apply PTQ to pre-trained LIC networks without retraining, 
and RDO-PTQ~\cite{ShiLM24} further introduces rate–distortion-aware block-wise reconstruction. 

From a broader perspective, existing LIC quantization methods still face a fundamental trade-off between efficiency, accuracy, and generality. Uniform quantization strategies, while simple, fail to reflect the highly non-uniform redundancy across network components, leading to inefficient precision allocation. In contrast, mixed-precision approaches developed for general vision tasks are often tailored to classification networks and rely on heuristic sensitivity metrics or expensive search procedures, making them difficult to transfer to LIC settings with rate–distortion optimization and entropy-constrained latent representations. Moreover, LIC models possess unique structural characteristics, such as hyperprior-based entropy modeling and iterative analysis–synthesis transforms, which further complicate sensitivity estimation and bit allocation. These differences highlight the necessity of a dedicated mixed-precision quantization strategy specifically designed for LIC.

To further improve efficiency and robustness, recent studies have shown that second-order information, particularly the Hessian matrix~\cite{DongYGMK19}, provides a principled measure of layer-wise sensitivity in neural networks. This insight enables more informed precision allocation by identifying modules that are more sensitive to quantization perturbations. Motivated by this observation, we propose a mixed-precision PTQ framework for LIC that enables efficient low-bit compression while ensuring numerical consistency across heterogeneous hardware platforms.

In our method, block-wise sensitivity is estimated using Hessian-based second-order information, and bit-width allocation is formulated as a constrained integer optimization problem under a target compression ratio. The optimization further incorporates rate–distortion-aware objectives to jointly minimize quantization error and task-specific distortion. The proposed framework, termed Hessian-Aware Mixed-Precision PTQ for LIC (HAMP-LIC), follows a four-stage pipeline (Fig.~\ref{fig:overview}): (1) sensitivity estimation, (2) task-aware metric construction, (3) mixed-precision allocation via constrained optimization, and (4) block-wise reconstruction refinement.

This article substantially extends our DCC 2026 paper, MPP-LIC~\cite{Zhang2026MPP}. 
Beyond the four-stage mixed-precision PTQ framework, it introduces task-aware sensitivity combining Hessian-trace information with rate–distortion degradation, 
formulates bit-width allocation as a global size-constrained optimization solved via Pareto-frontier search, 
and details block-wise reconstruction using sequential scaling optimization and adaptive rounding.
The main contributions of this work are summarized as follows:
\begin{enumerate}
\item We propose HAMP-LIC, a mixed-precision PTQ framework that integrates Hessian-based second-order sensitivity estimation with rate–distortion-aware optimization for LIC models.
\item We design a task-aware sensitivity refinement that modulates the Hessian-trace metric with the rate–distortion loss degradation of each block, so that bit-width decisions directly reflect the compression objective rather than generic reconstruction error.
\item We formulate bit allocation as a constrained integer optimization problem and introduce an efficient Pareto-frontier search strategy that reduces complexity from exponential to tractable scale, combined with progressive block-wise refinement.
\item We demonstrate that HAMP-LIC achieves up to 4.85$\times$ model compression with negligible BD-rate degradation, while ensuring cross-platform numerical consistency and eliminating floating-point-induced decoding mismatches.
\end{enumerate}

The remainder of this paper is organized as follows. Section~\ref{sec:related_work} reviews related work on neural network compression, quantization for LIC, and mixed-precision quantization. Section~\ref{sec:proposed_method} details the four steps of the proposed HAMP-LIC framework. Section~\ref{sec:experiments} presents the experimental settings, comparisons with state-of-the-art PTQ methods, and ablation studies. Finally, Section~\ref{sec:conclusion} concludes this paper.

\section{Related Work}
\label{sec:related_work}

In this section, we first review neural network compression methods in Section~\ref{sec:nn_compression}, followed by applications of model compression in LIC in Section~\ref{sec:lic_compression}. Finally, we discuss mixed-precision quantization methods in Section~\ref{sec:mixed_precision}.

\subsection{Neural Network Compression}
\label{sec:nn_compression}

Neural network compression aims to reduce computational cost and memory footprint while preserving model accuracy. Among existing techniques, quantization is particularly attractive due to its hardware efficiency, converting full-precision weights and activations into low-bit representations.

Quantization methods are typically categorized into Quantization-Aware Training (QAT) \cite{jacob2018quantization,jung2019learning} and Post-Training Quantization (PTQ)~\cite{banner2019post,frantar2022gptq}. QAT simulates quantization during training, allowing the model to adapt to quantization errors via gradient-based optimization~\cite{balle2019integer}. While effective under extremely low bit-widths, it requires full training data and incurs high computational cost. PTQ, in contrast, directly quantizes pre-trained models using a small calibration set to estimate activation statistics~\cite{He2022ptq}. Although efficient, PTQ often suffers from noticeable accuracy degradation in low-bit settings.

Recent reconstruction-based approaches aim to bridge this gap by optimizing quantization with unlabeled calibration data. AdaRound~\cite{Nagel2020UpOD} formulates weight rounding as a learnable optimization problem. BRECQ~\cite{li2021brecq} performs block-wise reconstruction to capture inter-layer dependencies. AdaQuant~\cite{Hubara2020improve} jointly optimizes weight rounding and activation clipping, while QDrop~\cite{WeiGLLY22} improves robustness by stochastically bypassing quantization during reconstruction. These methods operate at different granularities (e.g., tensor-, layer-, and block-level), balancing optimization flexibility and computational efficiency.

\subsection{Application of Model Compression in LIC}
\label{sec:lic_compression}

Model compression, particularly quantization, has become increasingly important in LIC to enable efficient deployment. Early work focused on mitigating the inconsistency of floating-point computation across hardware. Ballé et al.~\cite{balle2019integer} proposed integer-only LIC networks to ensure deterministic and hardware-friendly inference. Hong et al.~\cite{Hong0LPM21} further explored partial quantization by restricting fixed-point inference to the decoder, reducing complexity while preserving rate–distortion performance.

More recent efforts have shifted toward fully quantized LIC models. He et al.~\cite{He2022ptq} first applied Post-Training Quantization (PTQ) to pre-trained LIC networks, enabling efficient deployment without retraining, but suffering from performance degradation at low bit-widths. To address this, Sun et al.~\cite{SunYK25} reduced activation dynamic ranges via channel splitting and pruning, improving quantization robustness.

Recent advances incorporate reconstruction-based PTQ into LIC. Shi et al.~\cite{ShiLM24} integrated BRECQ~\cite{li2021brecq} for block-wise reconstruction and further introduced task-aware rate–distortion optimization, moving beyond MSE-based objectives. These efforts reflect a shift toward structured and task-aware quantization, narrowing the performance gap with full-precision LIC while maintaining efficiency.

\subsection{Mixed Precision Quantization}
\label{sec:mixed_precision}

Mixed-precision quantization exploits the observation that different layers in a neural network exhibit varying sensitivity to quantization~\cite{DongYGMK19,wang2019haq}. Instead of applying a uniform bit-width across all layers, it assigns higher precision to more sensitive layers and lower precision to less sensitive ones, thereby achieving a better trade-off between model size, computational efficiency, and accuracy. This adaptive allocation is particularly beneficial for deep architectures, where quantization errors can accumulate unevenly across layers.

Most existing approaches~\cite{wu2018mixed,ZhouMCF18} formulate mixed-precision quantization as a global optimization problem, aiming to determine the optimal bit-width configuration for each layer under constraints such as model size, latency, or energy consumption. These methods typically integrate bit-width selection into the training process, leveraging gradient-based optimization or reinforcement learning to jointly optimize network weights and precision assignments. While effective, such approaches often incur substantial computational overhead due to the large search space.

A key challenge arises from the combinatorial nature of the problem: the search space for layer-wise bit-width assignment grows exponentially with the number of layers, making exhaustive search intractable for modern deep networks. To alleviate this issue, sensitivity-based methods~\cite{DongYGMK19,HuangS0LHWXC22,YangJ21} have been proposed. These approaches first estimate the quantization sensitivity of each layer in a pre-trained model, typically based on metrics such as reconstruction error or loss perturbation. Bit-widths are then allocated by ranking layers according to their sensitivity, assigning higher precision to layers that are more critical to overall performance. This strategy significantly reduces the search complexity while maintaining competitive accuracy.

\section{Proposed Method}
\label{sec:proposed_method}

In this section, we present the proposed HAMP-LIC framework.
We first introduce the preliminaries (Section~\ref{sec:preliminaries}),
including the block-wise quantization formulation and the rate--distortion objective of LIC models, 
and then detail the four steps ((Sections~\ref{sec:step_1}–\ref{sec:step_4})) through which HAMP-LIC applies mixed-precision quantization to the pretrained LIC model.
The complete procedure is summarized in Algorithm~\ref{alg:hamp_lic}.

\subsection{Preliminaries}
\label{sec:preliminaries}

Assume that the LIC model can be partitioned into $L$ blocks denoted by 
$\{B_1, B_2, \ldots, B_L\}$, each associated with learnable weight parameters 
$\{W_1, W_2, \ldots, W_L\}$. Each block may consist of one or more layers, 
and the quantization of each block is performed independently to enable 
fine-grained reconstruction.

The $b$-bit uniform quantization operation applied to a weight parameter $W$ 
is defined as:
\begin{equation}
\begin{aligned}
W_b = \text{Clip}\left(\left\lfloor W / s \right\rceil, n, p\right)
\end{aligned}
\end{equation}
where $s$ denotes the quantization scale parameter that controls the mapping 
from floating-point values to integers, and $n$ and $p$ represent the negative 
and positive integer clipping thresholds, respectively, which are determined 
by the target bit-width $b$ as $n = -2^{b-1}$ and $p = 2^{b-1} - 1$ for 
signed quantization. The rounding operation $\lfloor \cdot \rceil$ maps each 
scaled value to its nearest integer, and the clipping operation $\text{Clip}(\cdot, \cdot)$ ensures that 
all values remain within the representable range. Following previous 
studies~\cite{ShiLM24, SunYK25}, we adopt per-channel scale quantization, 
where each output channel of a weight tensor is assigned an independent scale 
parameter $s$, allowing for finer adjustment of the quantization range and 
reducing the overall quantization error compared to per-tensor quantization.

For the LIC network, the overall training objective can be formulated as a 
rate-distortion loss function defined over the training set:
\begin{equation}
\begin{aligned}
\mathcal{L}(\theta) = \frac{1}{N} \sum_{i=1}^{N} J(x_i, \hat{x}_i; \theta)
\end{aligned}
\end{equation}
where $x \in X$ denotes the input image set, $\hat{x} \in \hat{X}$ 
denotes the corresponding reconstructed image set produced by the LIC model, 
and $N = |X|$ is the total number of training samples. The per-sample 
R-D loss $J(x_i, \hat{x}_i; \theta)$ combines a distortion term measuring
reconstruction fidelity (\emph{e.g.}, MSE or MS-SSIM) and a rate term penalizing 
the estimated bitrate of the latent representation, jointly encouraging the model 
to achieve a favorable rate-distortion trade-off.

\begin{algorithm}[t]
\caption{HAMP-LIC: Hessian-Aware Mixed-Precision PTQ for LIC}
\label{alg:hamp_lic}
\begin{algorithmic}[1]
\REQUIRE FP32 model $\mathcal{M}=\{B_l\}_{l=1}^{L}$; calibration set $\mathcal{X}_{\mathrm{cal}}$; bit-width set $B$; budget $\epsilon$.
\ENSURE Mixed-precision quantized model $\mathcal{M}_{mp}$.

\STATE \textit{// 1. Hessian-based sensitivity}
\FOR{$l=1$ to $L$}
    \STATE Estimate $\operatorname{Tr}(H_l)$ on $\mathcal{X}_{\mathrm{cal}}$ using Hutchinson's method;
\ENDFOR

\STATE \textit{// 2. Build task-oriented sensitivity list}
\STATE Evaluate the 8-bit reference loss $\mathcal{L}_{\mathrm{int8}}$ on $\mathcal{X}_{\mathrm{cal}}$;
\FOR{$l=1$ to $L$}
    \FOR{$b \in B$}
        \STATE Quantize $B_l$ to $b$ bits and evaluate $\mathcal{L}_{\mathrm{qp},l}(b)$;
        \STATE $\displaystyle \Omega_l(b) \gets \operatorname{Tr}(H_l)
        \frac{|\mathcal{L}_{\mathrm{qp},l}(b)-\mathcal{L}_{\mathrm{int8}}|}
        {|\mathcal{L}_{\mathrm{int8}}|}$;
    \ENDFOR
\ENDFOR

\STATE \textit{// 3. Mixed-precision allocation}
\STATE Rank blocks by task-aware sensitivity and enumerate monotone contiguous partitions as in Eq.~(\ref{eq:search_space_reduced});
\STATE $\displaystyle \{b_l^*\}_{l=1}^{L} \gets
\operatorname*{argmin}_{\{b_l\}} \sum_{l=1}^{L}\Omega_l(b_l)$
\STATE \hspace{2.1em}$\displaystyle \text{s.t.}\quad b_l\in B,\quad
\frac{1}{8L}\sum_{l=1}^{L}b_l\leq\epsilon$;

\STATE \textit{// 4. Block-wise reconstruction}
\FOR{$l=1$ to $L$}
    \STATE Set $\widehat{\boldsymbol{w}}^{\,l}=Q(\boldsymbol{w}^l;s_{\boldsymbol{w}}^l)$ and
    $\widehat{\boldsymbol{b}}^{\,l}=Q(\boldsymbol{b}^l;s_{\boldsymbol{b}}^l)$ using $b_l^*$ bits;
    \STATE Optimize $(s_{\boldsymbol{w}}^{l*},s_{\boldsymbol{b}}^{l*})$ by minimizing $(\widehat{J}-J_0)^2$;
    \STATE Optimize $\boldsymbol{V}^{l*}$ and the scale factors by minimizing
    $\lambda_t\mathcal{L}_{\mathrm{task}}+\mathcal{L}_{lq}$;
\ENDFOR
\RETURN $\mathcal{M}_{mp}$ with $\{b_l^*\}_{l=1}^{L}$, $\boldsymbol{s}^*$, and $\boldsymbol{V}^*$.
\end{algorithmic}
\end{algorithm}

\subsection{Step 1: Acquire Sensitivity List}
\label{sec:step_1}

Our approach begins with a full-precision pre-trained LIC network $\mathcal{M}$ and constructs 
a per-block sensitivity list to guide the subsequent mixed-precision quantization 
process. The sensitivity list characterizes how sensitive each block is to 
quantization-induced perturbations, enabling the allocation of higher bit-widths 
to more sensitive blocks and lower bit-widths to more robust ones.

A straightforward choice for measuring block sensitivity is the first-order 
gradient information. However, first-order metrics are often insufficient for 
accurately capturing layer-wise sensitivity, as they do not account for the 
curvature of the loss landscape~\cite{DongYGMK19}. To address this limitation, 
we adopt second-order information as the primary sensitivity metric. 
Specifically, we utilize the Hessian trace of the loss function with respect 
to the weights of each block, which reflects the local curvature and provides 
a more reliable indicator of how much the model output would change in response 
to weight perturbations introduced by quantization.

Formally, for block $B_l$ with weight parameters $W_l$, the Hessian matrix 
is defined as:
\begin{equation}
    H_l = \frac{\partial^2 \mathcal{L}}{\partial W_l^2}
\end{equation}
where $l \in \{1, \ldots, L\}$ and $\mathcal{L}$ is the loss function defined over the calibration set.
The Hessian trace $\text{Tr}(H_l)$ serves as the sensitivity score for block 
$B_l$, with a larger trace value indicating greater sensitivity to quantization\cite{DongYGMK19}.

Direct computation of the full Hessian matrix is computationally prohibitive 
for large-scale neural networks. To ensure computational tractability, we 
instead estimate the Hessian trace efficiently via Hessian-vector products 
combined with randomized probing techniques, based on Hutchinson's 
method~\cite{DongYGMK19,dong2020hawqv2}.

Once the Hessian trace is computed for all $L$ blocks, the sensitivity scores 
are collected and ranked to form the sensitivity list 
$\mathcal{S} = \{(B_l, \text{Tr}(H_l))\}_{l=1}^{L}$, 
which serves as the foundation for bit-width allocation in the subsequent step.

\subsection{Step 2: Build Task-Oriented Sensitivity List}
\label{sec:step_2}

While the Hessian trace obtained in Step 1 provides a 
geometry-aware measure of each block's sensitivity to weight perturbations, 
it does not directly reflect the impact of quantization on the downstream 
compression task. To address this, we introduce a task-constrained sensitivity 
metric that couples the second-order curvature information with the task-specific 
rate-distortion loss, enabling a more accurate assessment of each block's 
contribution to the overall model performance under quantization.

Specifically, for each block $B_l$ we define a task-aware sensitivity $\Omega_l$
as a composite measure that integrates both the potential sensitivity captured
by the Hessian trace and the task-oriented loss degradation induced by
quantization. Formally, $\Omega_l$ is expressed as:
\begin{equation}
\Omega_l(b_l) = \text{Tr}(H_l) \cdot
\frac{| \mathcal{L}_{\text{qp}}(b_l) - \mathcal{L}_{\text{int8}} |}
{| \mathcal{L}_{\text{int8}} |}
\label{eq:omega}
\end{equation}
where $L$ is the number of quantizable blocks, $b_l$ is the weight bit-width assigned to the $l$-th block, 
$\text{Tr}(H_l)$ is the Hessian trace of block $B_l$ computed in Step~1,
and $|\cdot|$ denotes the absolute value. 
The term $\mathcal{L}_{\text{qp}}(b_l)$ denotes the task loss
evaluated when the weights of block $B_l$ are quantized to $b_l$ bits, while
$\mathcal{L}_{\text{int8}}$ denotes the task loss of the reference model whose
weights are uniformly quantized to 8 bits. The ratio
$| \mathcal{L}_{\text{qp}}(b_l) - \mathcal{L}_{\text{int8}} | /
| \mathcal{L}_{\text{int8}} |$ measures the relative task loss degradation
introduced by quantizing block $B_l$, which serves as a task-aware weighting
factor to modulate the Hessian-based sensitivity. By combining these two terms,
$\Omega_l$ captures both the geometric sensitivity of each block and its
practical impact on the compression objective.

For the $i$-th sample of an LIC model with a hyperprior structure, the
rate-distortion (R-D) loss $J_i = J(x_i, \hat{x}_i; \theta)$ is defined as:
\begin{equation}
\begin{aligned}
J_i & = \lambda \cdot D_i + R_{\hat{\boldsymbol{Y}},i} + R_{\hat{\boldsymbol{Z}},i} \\
  & = \lambda \cdot D(x_i, \hat{x}_i) \\
  & + \mathbb{E}\!\left[-\log_2\!\left(p_{\hat{\boldsymbol{Y}}}
    (\hat{\boldsymbol{Y}}_i \mid \hat{\boldsymbol{Z}}_i)\right)\right]
  + \mathbb{E}\!\left[-\log_2\!\left(p_{\hat{\boldsymbol{Z}}}
    (\hat{\boldsymbol{Z}}_i)\right)\right],
\end{aligned}
\end{equation}
where $\lambda$ is the Lagrange multiplier that controls the trade-off between 
rate and distortion, $R_{\hat{\boldsymbol{Y}},i}$ is the estimated bitrate for
the quantized latent representation $\hat{\boldsymbol{Y}}_i$, and
$R_{\hat{\boldsymbol{Z}},i}$ is the estimated bitrate for the hyperprior latent
representation $\hat{\boldsymbol{Z}}_i$. $D(x_i, \hat{x}_i)$
denotes the distortion between the original image $x_i$ and its
reconstruction $\hat{x}_i$, which is typically measured by MSE or
MS-SSIM depending on the target application.

Given the task-aware sensitivities $\{\Omega_l\}$ defined in
Eq.~(\ref{eq:omega}), the quantization problem is formulated as a constrained
optimization that seeks the optimal per-block bit-width assignment
$\{b_l\}_{l=1}^{L}$ to minimize the accumulated sensitivity-weighted loss
degradation---an efficient proxy for the R-D loss $J$---while satisfying the
model compression ratio budget:
\begin{equation}
\begin{aligned}
&\operatorname*{argmin}_{\{b_l\}}\ \sum_{l=1}^{L} \Omega_l(b_l) \\
&\mathrm{s.t.}\ \frac{1}{8L}\sum_{l=1}^{L} b_l \leq \epsilon
\end{aligned}
\label{eqa:arg_min}
\end{equation}
where $\epsilon$ is a hyperparameter controlling the
overall compression ratio relative to 8-bit quantization, which is empirically
set to $\epsilon = 0.75$ in our experiments; a 10\% tolerance on the average
bit-width is allowed in practice when enumerating candidate allocations. This constraint ensures that the
average bit-width across all blocks does not exceed a prescribed budget, 
effectively limiting the model size overhead introduced by mixed-precision 
quantization. The detailed optimization procedure for solving this constrained
problem is described in Section~\ref{sec:step_3}.

\subsection{Step 3: Mixed-Precision Bit-Width Allocation via Pareto Frontier Search}
\label{sec:step_3}

Given the task-aware sensitivities $\{\Omega_l\}$ refined in Step~\ref{sec:step_2},
this step solves the constrained optimization problem in Eq.~(\ref{eqa:arg_min})
to determine the bit-width assignment for each block. Let $B$
denote the set of admissible bit-widths of size $k$ for each block, \emph{e.g.},
$B = \{2, 4, 8\}$, and the bit-width $b_l$ assigned to each block in
Eq.~(\ref{eqa:arg_min}) must satisfy $b_l \in B$. The set $\mathcal{B}$ subsequently
represents all feasible combinations of bit-widths across the entire network,
where each element corresponds to a complete assignment $\{b_1, b_2, \ldots, b_L\}$.

When formulated as a combinatorial search problem, the bit allocation space 
grows exponentially with the number of blocks, reaching $|\mathcal{B}| = k^L$ 
in the worst case. For example, with $B = \{2, 4, 8\}$ and a model with $L = 50$
quantizable blocks, the search space is $3^{50} \approx 7.12 \times 10^{23}$,
rendering exhaustive exploration computationally intractable.

To reduce the search space, we adopt a Pareto frontier approach guided by the 
sensitivity list. Blocks are first sorted in descending order of their sensitivity 
scores, and the search is reformulated as a partition problem over the sorted 
list, restricting assignments to monotonically non-increasing order with respect 
to sensitivity rank. The reduced search space is then expressed as:
\begin{equation}
|\mathcal{B}| = \sum_{j=1}^{k} \binom{k}{j} \cdot \binom{L-1}{j-1}
\label{eq:search_space_reduced}
\end{equation}
where $j$ indexes the number of distinct bit-width levels used, $\binom{k}{j}$
counts the ways to select $j$ bit-widths from $k$ candidates, and
$\binom{L-1}{j-1}$ counts the ways to partition $L$ sorted blocks into $j$
contiguous groups. This formulation reduces the search space from exponential
to polynomial in $L$, making the search computationally feasible. For the same
example ($L=50$, $k=3$), Eq.~(\ref{eq:search_space_reduced}) yields
$|\mathcal{B}| = 3 + 147 + 1176 = 1326$, a reduction of roughly 21 orders of
magnitude compared with the exhaustive search space of $3^{50} \approx 7.12
\times 10^{23}$. In our implementation with $k=3$, we enumerate the three-level
partitions of the sorted list (the $j=3$ terms), which dominate this reduced
space. Each candidate allocation is evaluated by the objective in
Eq.~(\ref{eqa:arg_min}), and the assignment that minimizes the accumulated
sensitivity while satisfying the model-size budget is selected as the final
bit-width configuration.

\subsection{Step 4: Task-Constraint Block-Wise Optimization}
\label{sec:step_4}

After acquiring the optimal bit-width allocation for each block in Step~3, to further 
improve the quantization performance, we adopt task-constraint block-wise optimization, 
which can be divided by optimization target into two parts: scaling optimization (Section~\ref{sec:scale_opt}) and rounding optimization (Section~\ref{sec:round_opt}).

\subsubsection{Scaling Optimization}
\label{sec:scale_opt}
We apply block-wise quantization to determine the optimal scale factors for weights and biases. The scale factors for the $l$-th block are obtained by minimizing the discrepancy between the quantized and full-precision rate-distortion loss:
\begin{equation}
s_{\boldsymbol{w}}^l, s_{\boldsymbol{b}}^l = \underset{s_{\boldsymbol{w}}^l,
s_{\boldsymbol{b}}^l}{\arg\min} \left( \widehat{J}\left(\hat{\boldsymbol{w}}^l,
\hat{\boldsymbol{b}}^l\right) - J_0\left(\boldsymbol{w}^l,
\boldsymbol{b}^l\right) \right)^2
\end{equation}
where $\widehat{J}(\hat{\boldsymbol{w}}^l, \hat{\boldsymbol{b}}^l)$ denotes the rate-distortion loss evaluated with the quantized weights $\hat{\boldsymbol{w}}^l$ and biases $\hat{\boldsymbol{b}}^l$ of the $l$-th block, $J_0(\boldsymbol{w}^l, \boldsymbol{b}^l)$ denotes the loss of the full-precision counterpart, and $s_{\boldsymbol{w}}^l$, $s_{\boldsymbol{b}}^l$ are the scale factors for the weights and biases of the $l$-th block, respectively. The optimization proceeds sequentially from the first block to the last. When optimizing the $l$-th block, all preceding blocks have already been quantized with fixed parameters, while all subsequent blocks retain their full-precision parameters.

\subsubsection{Rounding Optimization}
\label{sec:round_opt}

Rounding-to-nearest is a common default in PTQ but is suboptimal in general. Following AdaRound~\cite{Nagel2020UpOD}, AdaQuant~\cite{Hubara2020improve}, and BRECQ~\cite{li2021brecq}, we adopt an adaptive rounding strategy in which a learnable variable $\boldsymbol{V}$ is introduced to parameterize the rounding of weights, formally expressed as:
\begin{equation}
\underset{\boldsymbol{V}}{\arg\min} \left\| \Lambda\left(\boldsymbol{w}^l
\boldsymbol{x}^l\right) - \Lambda\left(\widehat{\boldsymbol{w}}^l
\boldsymbol{x}^l\right) \right\|^2 + \beta f_{\text{reg}}(\boldsymbol{V})
\end{equation}
where $\Lambda(\cdot)$ denotes the activation function and $\beta f_{\text{reg}}(\boldsymbol{V})$ is a differentiable regularization term, weighted by $\beta$, that encourages the rounded weights $\widehat{\boldsymbol{w}}$ to converge to integer values~\cite{Nagel2020UpOD}.

The per-block quantization loss is accordingly defined as:
\begin{equation}
\mathcal{L}_{lq} = \left\| \Lambda\left(\boldsymbol{w}^l \boldsymbol{x}^l\right)
- \Lambda\left(\widehat{\boldsymbol{w}}^l \widehat{\boldsymbol{x}}^l\right)
\right\|^2 + \beta f_{\text{reg}}(\boldsymbol{V})
\end{equation}
where $\widehat{\boldsymbol{x}}^l$ denotes the quantized input to the $l$-th block, accounting for the accumulated quantization error from preceding blocks.

The total optimization loss jointly combines the task-level loss and the block-wise quantization loss:
\begin{equation}
\boldsymbol{s}^*, \boldsymbol{V}^* = \underset{\boldsymbol{s}, \boldsymbol{V}}
{\arg\min}\ \lambda_t \mathcal{L}_{\text{task}} + \mathcal{L}_{lq}
\end{equation}
where $\boldsymbol{s}$ denotes the collection of scale factors for both weights and biases across all blocks, and $\lambda_t$ is a balancing hyper-parameter empirically set to $1$.
The resulting mixed-precision quantized model $\mathcal{M}_{mp}$ is obtained with the 
optimized scale factors $\boldsymbol{s}^*$, rounding variables $\boldsymbol{V}^*$, 
and per-block bit-width allocation $\{b_l\}_{l=1}^{L}$.

\begin{figure*}[t]
  \centering
  \includegraphics[width=0.49\linewidth]{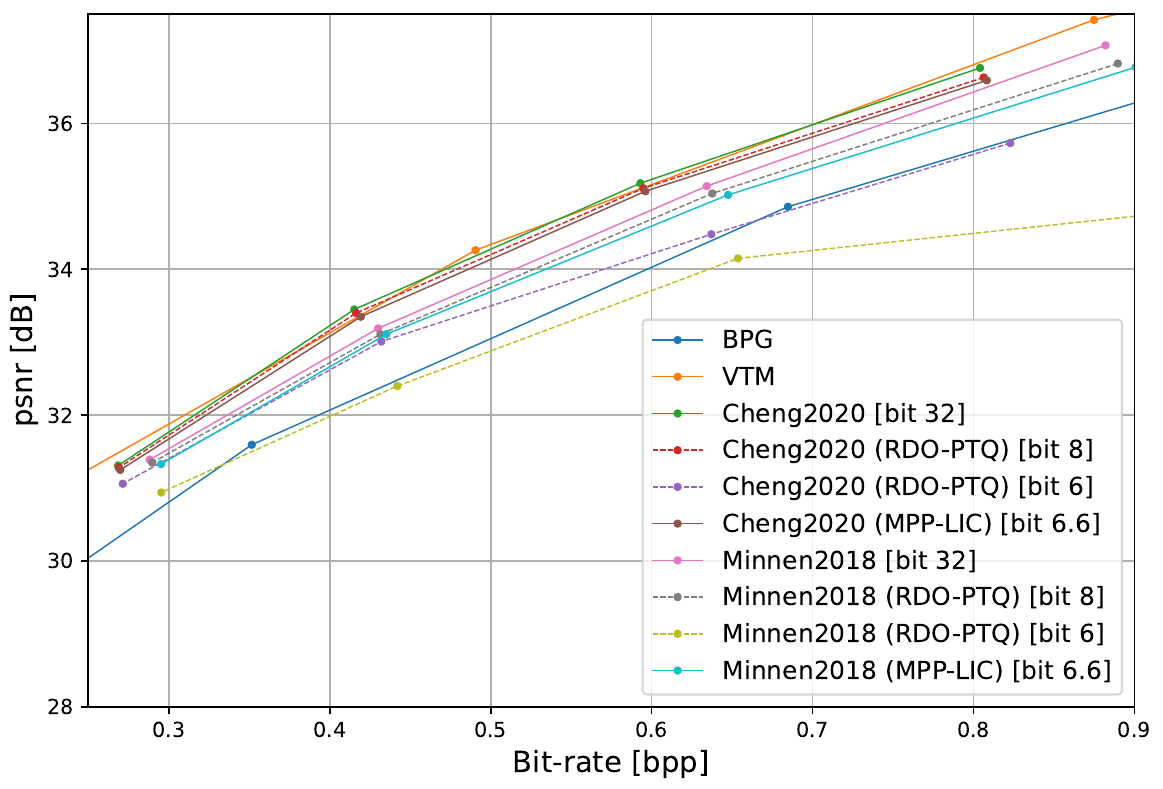}
  \includegraphics[width=0.49\linewidth]{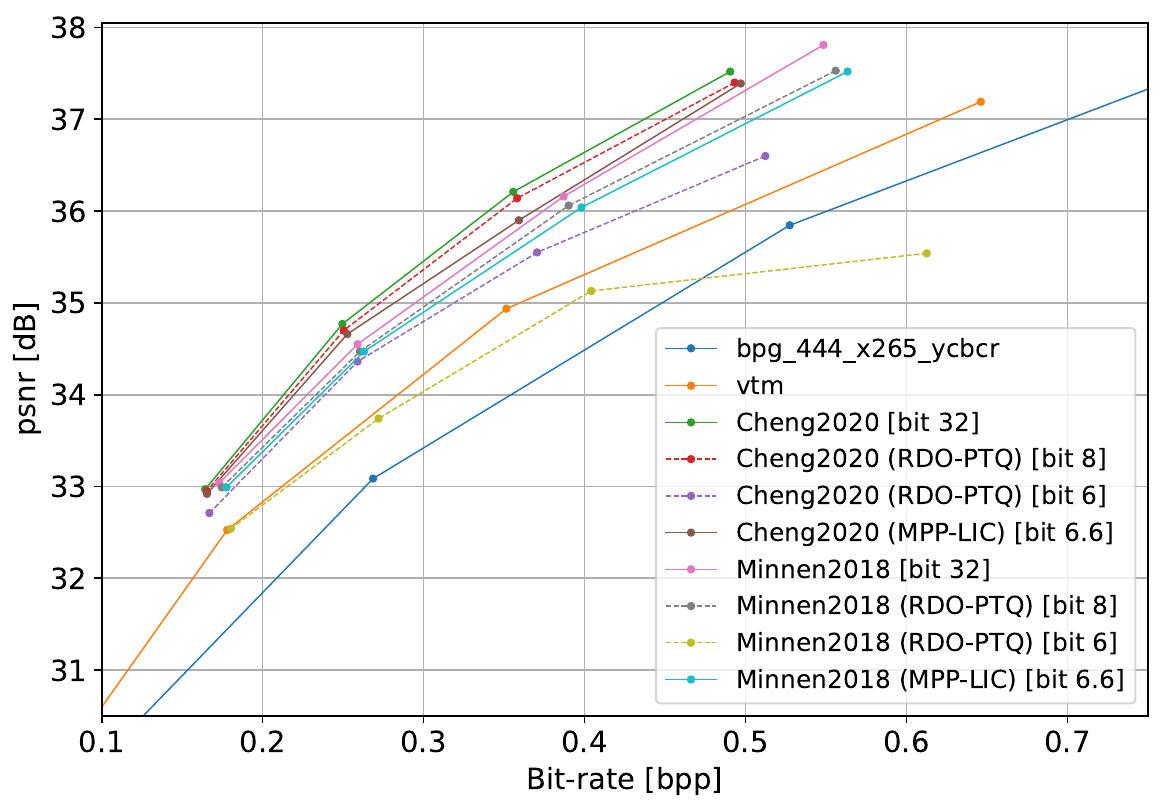}
  \caption{Compression performance evaluation on the Kodak dataset (left) and the CLIC dataset (right). Traditional codecs and unquantized LIC models are represented by solid lines, while quantized LIC models are shown with dashed lines, with quantization bits indicated in brackets [·]. 
}
  \label{fig:RD-curve}

\end{figure*}

\begin{figure*}[t]
\begin{center}
  \includegraphics[width=0.99\linewidth]{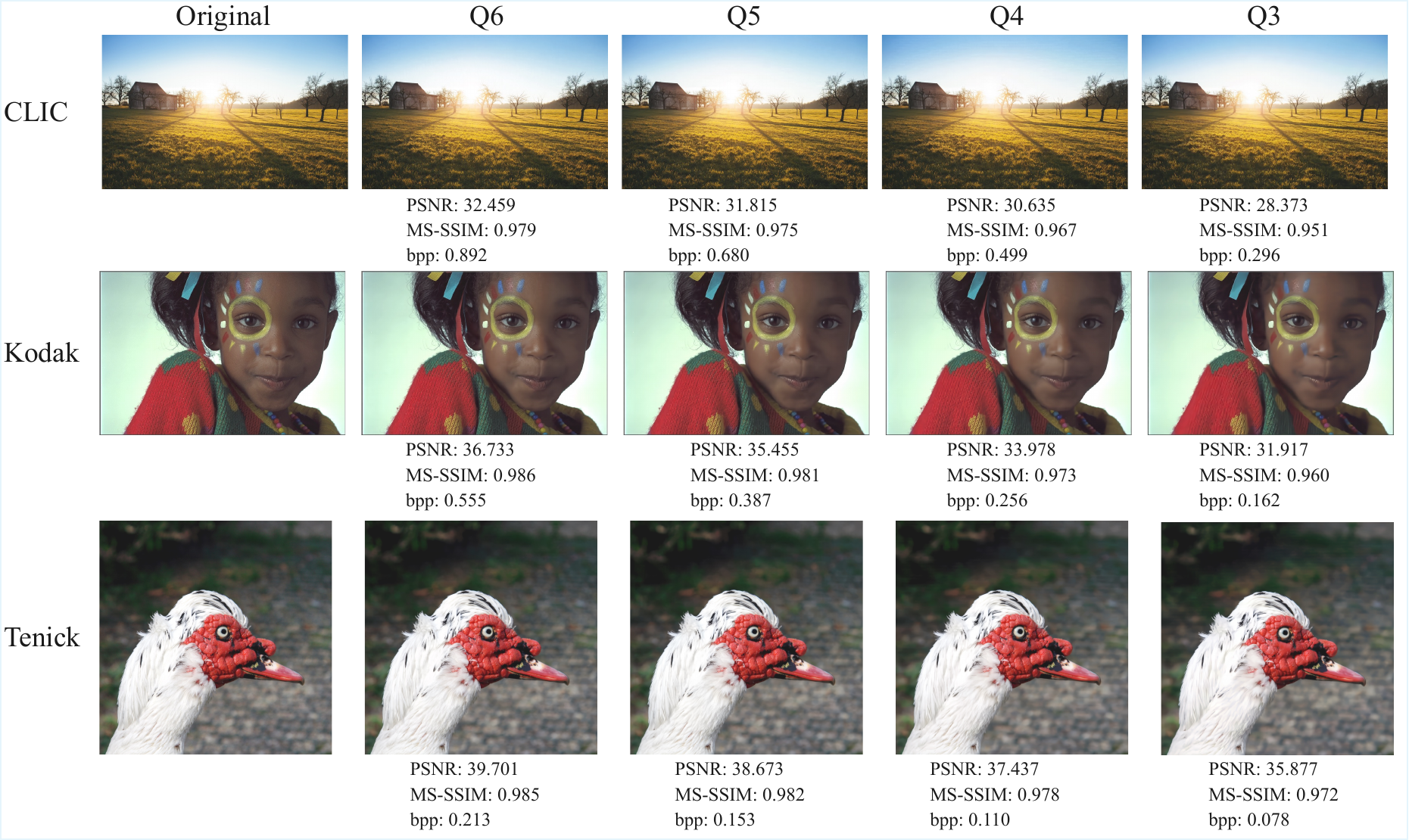}
\end{center}
\caption{Visual comparison of the proposed HAMP-LIC method across different quality levels (Q6 to Q3) on CLIC, Kodak, and Tecnick datasets. PSNR (dB), MS-SSIM, and bit-rate (bpp) are reported for each reconstruction.}
\label{fig:subjective_visual}
\end{figure*}

\section{Experiments}
\label{sec:experiments}
In this section, experimental settings and datasets are first described in Sections~\ref{sec:implementation_details}. Subsequently, we conduct experiments to justify the superiority of the proposed algorithm.

\subsection{Implementation Details}
\label{sec:implementation_details}

We first introduce the pretrained LIC models in Section~\ref{sec:pretrained_LIC}. Next, we display training details of the proposed HAMP-LIC in Section~\ref{sec:training_setup}. Then, comparison methods and datasets are described in Section~\ref{sec:methods_datasets}.

\subsubsection{Pretrained FP32 LICs}
\label{sec:pretrained_LIC}

Two representative LIC models are selected for evaluation: \textit{Minnen2018}~\cite{minnen2018joint} and \textit{Cheng2020}~\cite{cheng2020image}. Both models are based on a hierarchical variational autoencoder (VAE) architecture. Minnen2018 introduced an autoregressive context model that, for the first time, brought LIC performance to a level comparable with the traditional image codec BPG/H.265. Cheng2020 replaced the single Gaussian prior with a discretized Gaussian mixture model, achieving performance comparable with VVC/H.266.

Full-precision pre-trained models are obtained from the CompressAI library~\cite{Bgaint2020CompressAIAP}. Both models are trained using MSE as the distortion metric in the R-D loss, formulated as $\lambda \cdot 255^2 \cdot D + R$, where the Lagrange multiplier $\lambda$ is selected from $\{0.0067, 0.013, 0.025, 0.0483\}$, corresponding to four different rate-distortion operating points.

\subsubsection{Training Setup}
\label{sec:training_setup}

Training is performed with a calibration dataset, which only consists of 12 images randomly 
selected from the CLIC dataset. The proposed framework is optimized using the ADAM \cite{kingma2014adam}  
optimizer with a learning rate of 0.001. The mini-batch size is 4, and the input image during training is cropped to 256 × 256. 
The optimization iterations for each block are 20,000. For the Hessian-trace estimation in Step~\ref{sec:step_1}, we adopt Hutchinson's method with 5 Rademacher random vectors per block over 12 calibration samples (mini-batch size 4).

\subsubsection{Comparison Methods and Datasets}
\label{sec:methods_datasets}

For quantization methods, we make comparisons with both fixed-precision and mixed-precision PTQ methods.
The fixed-precision baselines include Range-Adaptive Quantization (RAQ)~\cite{Hong0LPM21},
task-oriented PTQ (RDO-PTQ)~\cite{ShiLM24},
and the fixed-precision variant (FPQ) reported in~\cite{FaisalHossain2024FlexibleMP}.
The mixed-precision baseline is Flexible Mixed Precision Quantization (FMPQ)~\cite{FaisalHossain2024FlexibleMP}.
Among these, RDO-PTQ and FMPQ are specifically designed for LIC models:
RDO-PTQ is proposed for fixed-point PTQ, while FMPQ targets mixed-precision PTQ
but is limited to an 8-bit weight quantization target.
The proposed HAMP-LIC overcomes this limitation and achieves superior performance.

For traditional image coders, we compare our results against the H.265/HEVC and H.266/VVC standards, using BPG software \cite{bellard2014bpg} and the VTM 23.11 test software \cite{VTM}, respectively. Both codecs were configured in all-intra mode with 8-bit YCbCr 4:4:4.

Three benchmark datasets are used for evaluation: Kodak~\cite{kodak}, Tecnick~\cite{tecnick}, and CLIC~\cite{clic2020}. The Kodak dataset consists of 24 uncompressed images at $768 \times 512$ resolution. The Tecnick dataset contains 100 high-resolution images at $1200 \times 1200$. The CLIC dataset refers to the professional validation set of the CLIC 2020 challenge, comprising 41 high-quality images with diverse content and resolutions.

\begin{table*}[ht]
\renewcommand\arraystretch{0.9}
\setlength\tabcolsep{10pt}
\small
\begin{center}
\begin{tabular}{@{}lllll@{}}
\toprule
\multirow{2}{*}{Model} & \multirow{2}{*}{Method} & \multicolumn{2}{l}{BD-Rate Loss (\%)} & \multirow{2}{*}{\begin{tabular}[c]{@{}l@{}}Compression \\ Ratio\end{tabular}} \\ \cmidrule(lr){3-4}
 &  & Kodak & Tecnick &  \\ \midrule
\multirow{6}{*}{Cheng2020} & RAQ & 27.84 & 29.95 & 4x \\
 & RDO-PTQ & 2.95 & 5.09 & 4x \\
 & FPQ & 2.05 & 4.97 & 4x \\
 & FMPQ & 0.89 & 2.68 & 3.99x \\
 & \begin{tabular}[c]{@{}l@{}}Proposed HAMP-LIC\\ (w=6.6, a=32)\end{tabular} & 0.59 & 1.79 & 4.85x \\
 & \begin{tabular}[c]{@{}l@{}}Proposed HAMP-LIC\\ (w=6.6, a=8)\end{tabular} & 2.99 & 6.33 & 4.85x \\ \midrule
\multirow{6}{*}{Minnen2018} & RAQ & 30.41 & 31.55 & 4x \\
 & RDO-PTQ & 1.19 & 4.01 & 4x \\
 & FPQ & 3.54 & 5.78 & 4x \\
 & FMPQ & 1.2 & 2.64 & 3.97x \\
 & \begin{tabular}[c]{@{}l@{}}Proposed HAMP-LIC\\ (w=6.6, a=32)\end{tabular} & 1.23 & 3.75 & 4.85x \\
 & \begin{tabular}[c]{@{}l@{}}Proposed HAMP-LIC\\ (w=6.6, a=8)\end{tabular} & 4.93 & 5.08 & 4.85x \\ \bottomrule
\end{tabular}
\end{center}

\caption{BD-Rate loss of various quantization methods relative to the original full-precision (FP32) LIC model. The ``x'' denotes the compression ratio of the weights; for instance, ``4x'' corresponds to quantization from 32-bit floating-point to 8-bit integer precision.
}
\label{table:rd_rate}
\end{table*}

\begin{table*}[ht]
\renewcommand\arraystretch{0.9}
\setlength\tabcolsep{8pt}
\small
\begin{center}
\begin{tabular}{@{}lllll@{}}
\toprule
Model                 & \multicolumn{2}{c}{FP32 Model} & \multicolumn{2}{c}{Proposed HAMP-LIC} \\ \midrule
Encode/Decode Device  & CPU/GPU        & GPU/CPU       & CPU/GPU           & GPU/CPU          \\ \midrule
Error Rate on Kodak   & 24/24          & 21/24         & 0/24              & 0/24             \\
Error Rate on Tecnick & 98/100         & 97/100        & 0/100             & 0/100            \\ \bottomrule
\end{tabular}
\end{center}

\caption{Decoding error rates of \textit{Cheng2020} during inference with and without PTQ.
The encoding process is executed on one platform (CPU or GPU), whereas decoding is conducted on the other (GPU or CPU), where CPU/GPU denotes encoding on the CPU and decoding on the GPU.}
\label{table:error_rate}
\end{table*}

\begin{figure}[t]
\begin{center}
  \includegraphics[width=0.95\linewidth]{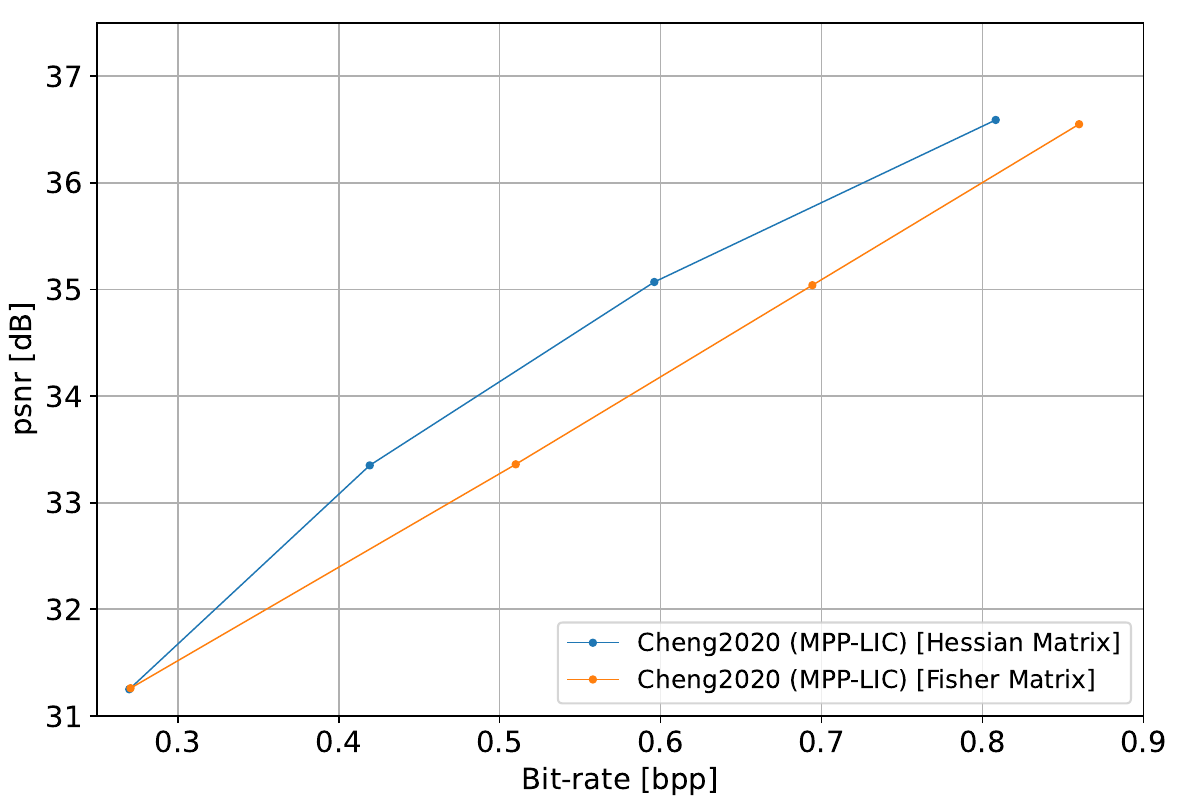}
\end{center}

\caption{Comparison of Hessian trace with Fisher matrix as the sensitivity metric used in Step 1.}
\label{fig:fisher_vs_hessian}
\end{figure}

\subsection{Evaluation}

In this section, we present the experimental results of the proposed HAMP-LIC method, including R-D performance (Section~\ref{sec:R_D_performance}), BD-rate comparison (Section~\ref{sec:BD_rate_comparison}), subjective visual comparison (Section~\ref{sec:subjective_visual_comparison}), and encoding/decoding error rate (Section~\ref{sec:enc_dec_error_rate}).

\subsubsection{R-D Performance}
\label{sec:R_D_performance}

We evaluate the rate-distortion (R-D) efficiency by plotting PSNR (dB) against bit-rate (bpp) in Fig.~\ref{fig:RD-curve}. The comparison includes traditional codecs, full-precision anchors, and PTQ-quantized LIC models across the Kodak and CLIC datasets.

The R-D curves demonstrate that our proposed HAMP-LIC method closely aligns with the floating-point models, showing negligible performance degradation even at ultra-low bit-rates. 
Compared to uniform PTQ and prior RDO-based approaches, HAMP-LIC achieves a superior R-D frontier, 
effectively narrowing the gap between quantized and full-precision performance. Furthermore, our method remains competitive with the VVC (VTM) standard. 
These results confirm that the mixed-precision strategy successfully preserves the model's representation power while significantly enhancing computational efficiency for practical deployment.

\subsubsection{BD-Rate Comparison}
\label{sec:BD_rate_comparison}
The proposed method is evaluated in comparison with existing PTQ approaches, using BD-rate loss relative to the full-precision models on two benchmark datasets, namely Kodak and Tecnick.
As shown in Table~\ref{table:rd_rate}, the proposed HAMP-LIC achieves the highest model compression ratio (4.85×) among all compared methods. With full-precision activations (w=6.6, a=32), it also attains the lowest BD-rate loss on the \textit{Cheng2020} model (0.59\% on Kodak and 1.79\% on Tecnick), outperforming the mixed-precision FMPQ baseline (0.89\% and 2.68\%) despite the more aggressive weight compression, and it remains competitive on the \textit{Minnen2018} model.

We also evaluate the proposed HAMP-LIC method with varying activation quantization bit widths. While quantizing activations does not reduce model size, it does impact compression performance. When reducing activation precision from full FP32 to 8-bit, the BD-rate loss increases by 2.40\% and 4.54\% for the \textit{Cheng2020} model on the Kodak and Tecnick datasets, respectively, and by 3.70\% and 1.33\% for the \textit{Minnen2018} model. The resulting fully quantized configuration (w=6.6, a=8) trades this modest BD-rate loss for reduced computation (see the BOPS analysis in Section~\ref{sec:bops_analysis}) and deterministic cross-platform decoding (Section~\ref{sec:enc_dec_error_rate}).

These results indicate that HAMP-LIC preserves reconstruction quality at a substantially higher weight compression ratio than alternative quantization techniques such as RDO-PTQ and prior mixed-precision methods.

\subsubsection{Subjective Visual Comparison}
\label{sec:subjective_visual_comparison}

In Fig.~\ref{fig:subjective_visual}, we present the subjective visual comparison across the CLIC, Kodak, and Tecnick datasets. The visualizations compare Original uncompressed images against reconstructions at decreasing quality levels from Q6 to Q3.

The proposed method preserves complex textures and color fidelity exceptionally well at ultra-low bit allocations. As quality decreases to Q3 , bitrates drop significantly to 0.296 bpp , 0.162 bpp , and 0.078 bpp. Despite extreme compression, visual degradation is minimal, avoiding severe artifacts like blocking or over-smoothing. This high perceptual quality is supported by strong MS-SSIM scores of 0.951 , 0.960 , and 0.972 at Q3, directly corroborating the superior compression efficiency demonstrated in our BD-rate analysis.

\subsubsection{Encoding/Decoding Error Rate}
\label{sec:enc_dec_error_rate}

As shown in Table~\ref{table:error_rate}, the FP32 model suffers from significant decoding errors when encoding and decoding are performed on different platforms (CPU/GPU or GPU/CPU). This indicates a severe lack of robustness and platform independence, likely caused by numerical non-determinism between the different hardware architectures (e.g., subtle differences in floating-point computation on CPUs vs. GPUs) \cite{balle2019integer, He2022ptq}.
In contrast, the Proposed HAMP-LIC method reduces the error rate to zero on both datasets and for both cross-platform scenarios, ensuring reliable decoding regardless of the hardware used for encoding or decoding.

\subsection{Ablation Study}
\subsubsection{Sensitivity Metric Analysis}

To further investigate the efficacy of various sensitivity measures, we conduct experiments using the Fisher information matrix as a substitute for the Hessian matrix.
As presented in Fig.~\ref{fig:fisher_vs_hessian}, it reveals that utilizing the Hessian matrix yields superior results compared to the Fisher matrix under the same bit-rate restriction.
Second-order metrics like the Hessian are superior to first-order gradients because they capture the curvature of the loss landscape, not just its slope \cite{DongYGMK19}. This provides a more accurate model of each parameter's sensitivity, leading to more effective optimization and higher-performance quantization.


\begin{figure*}[t]
\begin{center}
  \includegraphics[width=0.9\linewidth]{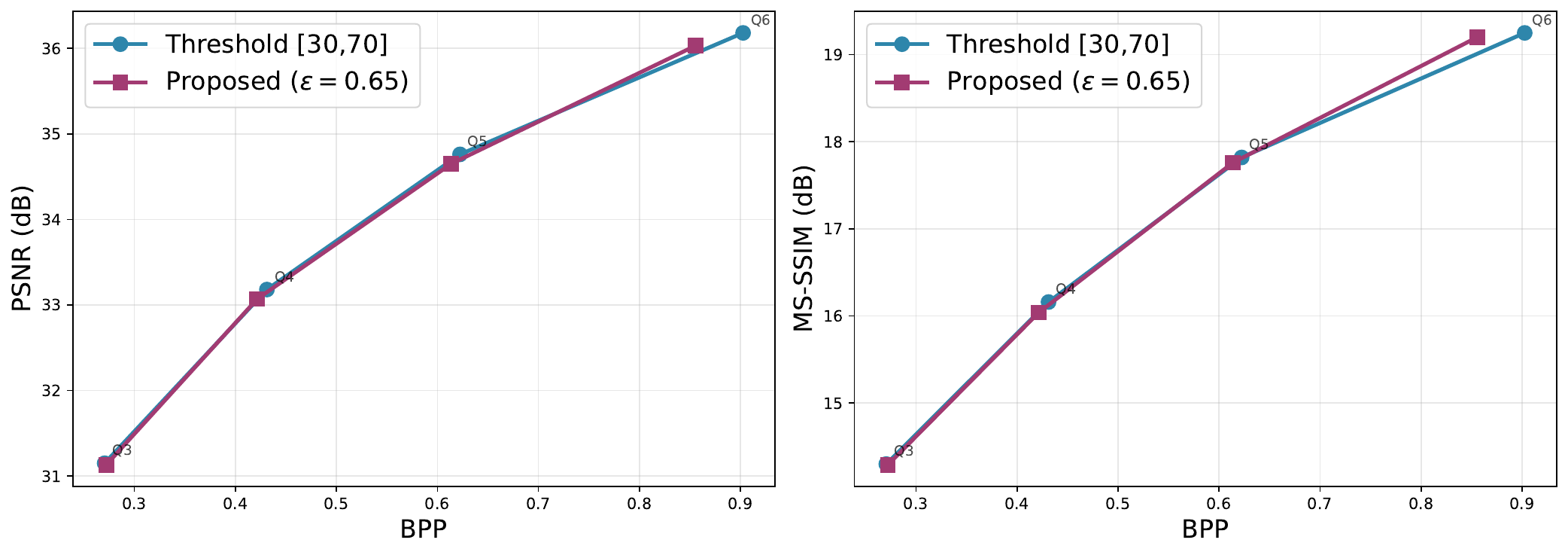}
\end{center}
\caption{Comparison of rate distortion performance between the proposed method ($\epsilon=0.65$, average bit-width of 5.719) and the threshold-based baseline (boundaries at 30\% and 70\%, average bit-width of 6).}
\label{fig:bit_allocation_comparison}
\end{figure*}

\begin{figure}[t]
\begin{center}
  \includegraphics[width=0.95\linewidth]{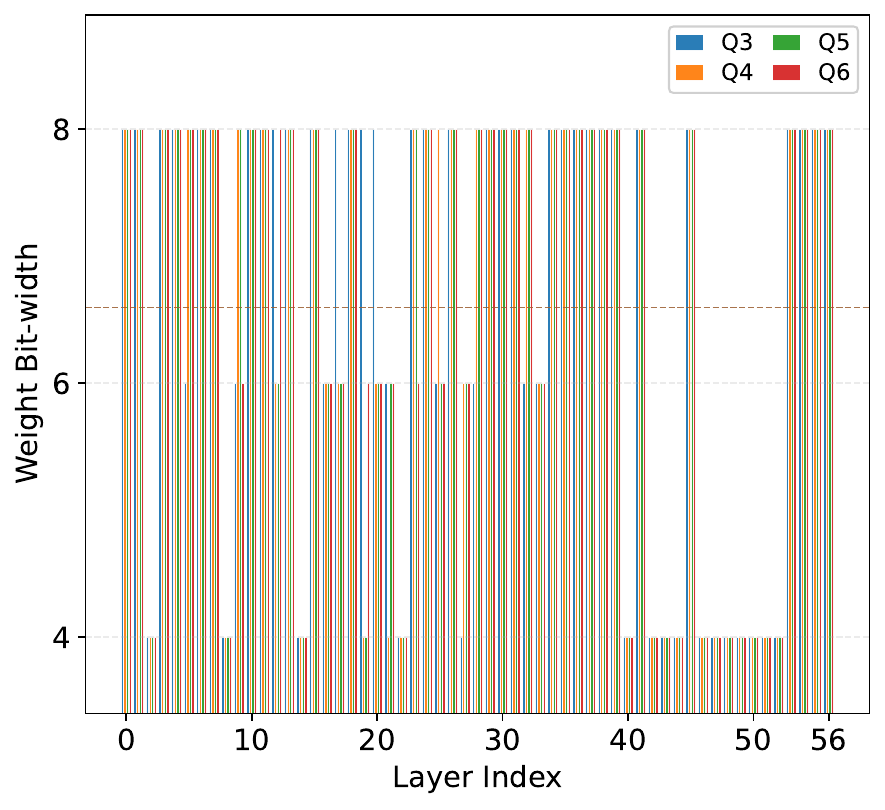}
\end{center}
\caption{Bit-width distribution across the \textit{Cheng2020} model layers for four different quality levels.}
\label{fig:lambda_layer_bits_grouped}
\end{figure}

\begin{table}[t]
\setlength\tabcolsep{2pt}
\centering
\caption{BOPS comparison across quality levels (input resolution $512\times768$, $b^a=8$).}
\label{tab:bops}
\begin{tabular}{lccccc}
\toprule
Quality & $\lambda$ & MACs (G) & \makecell{BOPS-Mixed\\(G)} & \makecell{BOPS-FP32\\(G)} & \makecell{vs.\ W8A8} \\
\midrule
Q3 & 0.0067 & 34.3 & 2007.5 & 35138.5 & $\downarrow$8.6\% \\
Q4 & 0.0250 & 76.9 & 4628.2 & 78780.0 & $\downarrow$6.0\% \\
Q5 & 0.0250 & 76.9 & 4652.6 & 78780.0 & $\downarrow$5.5\% \\
Q6 & 0.0483 & 76.9 & 4661.9 & 78780.0 & $\downarrow$5.3\% \\
\bottomrule
\end{tabular}
\end{table}

\subsubsection{Bit Allocation Methods}

To evaluate the bit allocation strategy from Step~\ref{sec:step_3}, we compare the proposed optimization in Equation~\ref{eqa:arg_min} against a baseline threshold approach. In the baseline, bits are assigned by thresholding the sensitivity list: the most sensitive 30 percent of blocks receive 8 bits, the middle 40 percent receive 6 bits, and the least sensitive 30 percent receive 4 bits, yielding an average bit-width of 6 bits. Conversely, the proposed method achieves a lower average bit-width of 5.719 with $\epsilon$ set to 0.65.
Experimental results in Figure~\ref{fig:bit_allocation_comparison} show that the proposed method maintains rate distortion performance consistent with the baseline, even outperforming it at higher bitrates (bpp $>$ 0.6). These findings demonstrate that the proposed optimization effectively reduces average bit weight while preserving reconstruction quality.

\subsubsection{Bit Distribution for Each Layer}

We conduct a study of the bit-width distribution across the \textit{Cheng2020} model quantized via our HAMP-LIC method. 
In Fig.~\ref{fig:lambda_layer_bits_grouped}, we plot these distributions for four quality levels.
The \textit{Cheng2020} model contains 57 quantizable layers: layers [0--18] comprise the main encoder ($g_a$), [19--40] the main decoder ($g_s$), [41--45] the hyper-encoder ($h_a$), [46--52] the hyper-decoder ($h_s$), and [53--56] the entropy model (entropy parameters and context prediction).
We observe that main-path weights (layers [0--40]) are generally more sensitive to quantization, requiring higher bit-widths than hyper-path weights. Specifically, the last layers of the encoder (layer 18) and decoder (layer 39) necessitate high bit-widths, as they encode critical information regarding the latent representation and the reconstructed output image, respectively.

\subsubsection{BOPS Analysis}
\label{sec:bops_analysis}

To further quantify the computational savings of our FMPQ method, we evaluate the Bit Operations (BOPS) for each quality level, defined as:
\begin{equation}
    \text{BOPS} = \sum_{i} \text{MACs}_i \times b^w_i \times b^a,
\end{equation}
where $\text{MACs}_i$ and $b^w_i$ denote the multiply-accumulate operations and weight bit-width of layer $i$, respectively, and $b^a = 8$ is the fixed activation bit-width. As reported in Table~\ref{tab:bops}, our mixed-precision quantization achieves BOPS of 2007.5\,G, 4628.2\,G, 4652.6\,G, and 4661.9\,G for quality levels $\lambda \in \{0.0067, 0.025, 0.025, 0.0483\}$, 
corresponding to reductions of 94.3\%, 94.1\%, 94.1\%, and 94.1\% relative to the full-precision (FP32) baseline. 
Compared to uniform 8-bit weight quantization (W8A8), our method further reduces BOPS by 8.6\%, 6.0\%, 5.5\%, and 5.3\%, respectively. 
This gain stems directly from the bit-width allocation described above: the hyper-path layers ($h_a$ and $h_s$, layers [41--52]), which are identified as less sensitive by our Hessian-based metric, are predominantly assigned 4-bit weights, whereas W8A8 applies a fixed 8-bit budget uniformly. The main-path layers ($g_a$ and $g_s$, layers [0--40]) retain higher bit-widths to preserve rate-distortion performance, contributing the majority of total BOPS. These results confirm that our sensitivity-aware bit allocation not only maintains reconstruction quality but also yields a measurable reduction in computational cost beyond that achieved by simple uniform low-bit quantization.

\section{Conclusion}
\label{sec:conclusion}
In this paper, we presented HAMP-LIC, a Hessian-aware mixed-precision post-training quantization framework for deploying learned image compression models on resource-constrained hardware. 
HAMP-LIC combines Hessian-trace-based block sensitivity with a task-aware rate–distortion criterion, allocates bit widths through an efficient Pareto-frontier search, and applies block-wise reconstruction to reduce quantization error. 
Experiments on the \textit{Minnen2018} and \textit{Cheng2020} models across multiple benchmark datasets show that HAMP-LIC achieves up to 4.85$\times$ model compression with negligible BD-rate loss, outperforms existing fixed- and mixed-precision PTQ methods, and completely eliminates cross-platform encoding–decoding errors. Future work will extend HAMP-LIC to transformer-based LIC architectures, automate the selection of the compression-ratio hyperparameter $\epsilon$, and reduce calibration overhead.
\bibliographystyle{IEEEtran}
\bibliography{refs}

\newpage

 




\vfill

\end{document}